# An Improved Fatigue Detection System Based on Behavioral Characteristics of Driver

Rajat Gupta
Department of Electronics Engineering
Indian Institute of Technology (ISM)
Dhanbad, India
e-mail: rajatstrc@gmail.com

Kanishk Aman
Department of Electronics Engineering
Indian Institute of Technology (ISM)
Dhanbad, India
e-mail: kanishkaman219@gmail.com

Nalin Shiva
Department of Applied Physics
Indian Institute of Technology (ISM)
Dhanbad, India
e-mail: nalinshiva81@gmail.com

Yadvendra Singh
Department of Electronics Engineering
Indian Institute of Technology (ISM)
Dhanbad, India
e-mail: yad3786@gmail.com

**Abstract—In recent years, road accidents have increased significantly. One of the major reasons for these accidents, as reported is driver fatigue. Due to continuous and longtime driving, the driver gets exhausted and drowsy which may lead to an accident. Therefore, there is a need for a system to measure the fatigue level of driver and alert him when he/she feels drowsy to avoid accidents. Thus, we propose a system which comprises of a camera installed on the car dashboard. The camera detect the driver's face and observe the alteration in its facial features and uses these features to observe the fatigue level. Facial features include eyes and mouth. Principle Component Analysis is thus implemented to reduce the features while minimizing the amount of information lost. The parameters thus obtained are processed through Support Vector Classifier for classifying the fatigue level. After that classifier output is sent to the alert unit.**

## I. INTRODUCTION

According to WHO, about 1.25 million people die globally each year due to road accidents[1]. This figure is errored due to the under-reporting. Fatigue driving is one of the key causes of these accidents[2]. India is one of the largest contributors to this number. Today there are many numbers of technologies developed for fatigue monitoring [3]-[16]. The drowsy state detection system can be classified into three kinds. The first one is based on an attribute such as steering wheel movement, lane position, acceleration, distance to the nearby vehicles, etc. But these type of system is constrained bylimitation like road state, away of driving, the vehicle used, etc. In the second kind of system, a physiological signal such as electroencephalogram (EEG), electromyogram (EMG), electrooculogram (EOG), electrocardiogram(ECG) is used to detect the fatigue level. Physiological signal based system is the most promising fatigue detection system but they require sensor attached to the skin which may affect the user by causing skin irritation, revulsion, loathing, repulsion, etc. The third kind of system uses characteristics like eye blinking, yawning, head pose, etc. to monitor the behavior of the driver and alert the driver if any of drowsiness symptoms are detected. Based on this three kind of systems and their fusions there are several types of products are commercially available in the market. But some of them make an alarm when the driver maybe goes to the microsleep and alarm wakes up the driver and maybe become the cause of abrupt reaction of the driver which may lead to an accident.Many other possible approaches which are subject dependent and require calibration for proper working [17]. In this paper, we propose a more practical, subject independent, robust, calibration free, behavioral based system.

## II. PROPOSED APPROACH

Research has helped to identify some signs or symptoms which help in determining the drowsy state of the driver[18]-[19].

These signs or symptoms are the following:
- Daydreaming and lack of focusing.
- Blinking frequently and partially closed eye.
- Not able to remember the traveled path.
- Yawning after every small period.
- Drifting or maybe move out from the lane.
- Head nodding.
- Poor Concentration
- Slow reactions.

Here we made our focus on the features related to eye and mouth like frequent blinking, yawning etc., for classifying the fatigue level of a driver. The detection system of fatigue, as proposed by us is being divided into four parts: First, a video camera is used for the live streaming of the driver while driving and sends the video feed to a computer vision system that can detect the driver's face in the video frame. After getting the face image it sends to Support Vector Machine (SVM) classifier which classifies the facial image as fatigued or not fatigued. The output of the classifier is represented +1 for fatigued and -1 for not fatigued and this number is input to a running sum that adds the consecutive output values. These outputs then serve to be the input for alert system unit which then acts upon these inputs to produce a resultant of them. This resultant with respect to time is then classified into different fatigue level i.e. no, low and high fatigue level. For different fatigue level, the action taken by the alert unit is different which is explained in the later part.

### III. DATA

Generally, we undermine the importance of data. But data is equally or we can say not less important than the algorithm that uses the dataset for training. Large data can outrun a better algorithm and more the data is versatile more it will be better. In our case, the dataset should contain the wide range of facial images viz. talking faces, smiling faces, face with both dark and transparent glasses. In the case of dark glasses, our system learns to works only on the mouth feature set. Also, images of the people who put a hand on the mouth while yawning help our system to predict correctly for the fatigue. Images with dim light and broad light also need to be considered.

### IV. PROPOSED SYSTEM

The proposed system is divided into four the following subsystems:

#### A. Video Capture Unit

The video capture unit used to record the video in real time of the frame containing the driver face through a camera placed on the car dashboard. The video is sampled with some frequency and the sampled frame is sent to the face detection unit.

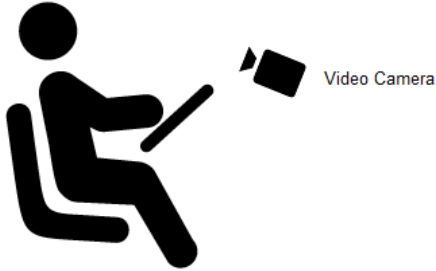

Figure 1. Camera as used in the video capture unit

#### B. Face Detection Unit and Features Extraction

This unit receives the sampled video frame from the video capture unit. The images from the video capture unit are the RGB image and for the very dim light condition, we perform low-light image enhancement and noise elimination [20]. For improving the accuracy of our system, we eliminate the noise of the image before amplifying it through contrast enhancement. The above process is divided into two subtasks. At first, for de-noising the image we apply the superpixel based adaptive noising and secondly, for amplifying the image luminance adaptive contrast enhancement is used. We need to de-noise the image before contrast enhancement, so that the noise has been eliminated before its amplification through contrast enhancement. The above method increases the accuracy of our system significantly as it eliminates heavy noise, texture blurring and over-enhancement from the image which is then processed as accordingly. The image is changed to the grayscale image because for face detection we do not need the color data. For face detection in the frame, we use the rapid object detection which uses the boosted cascade of the classifier by Viola-Jones that works with the Haar-like features [21]. The face detection method returns the abscissa, ordinates, length, and breadth of the rectangle boxed in the facial image.

As we need only the eyes and mouth part of the face for the feature set. For this first, we scaled the rectangle boxed in face image to 100*100 pixels and then we use the 80x30 rectangular pixels window for eye and 40x40 rectangular pixels window for the mouth. The x- and y-coordinates of the rectangular window is (10, 20) and (30, 60) for the eye and mouth respectively. The above part is then transferred to the fatigue detection unit for further processing.

#### C. Fatigue Detection on Extracted Features

From the face detection unit, we get a sequence of eye and mouth image of the driver. Now from this extracted dataset, we can perform fatigue detection analysis on various facial features. These facial features include eyes (fast blinking or heavy eyes), mouth (yawn detection). The combined result of fatigue detection on these facial features is used to give the final result as to whether the driver is in fatigue or alert state.

The images from face detection unit, in pixels, can contain a lot of features. If we perform as said in the previous unit we would get a feature vector of size 4000. This size can be further reduced with the use of Principle Component Analysis (PCA). PCA can be used to avoid the problem caused by high-dimensionality as it compresses the data while minimizing the amount of information lost. It searches for a pattern in the data and reduces as much possibly correlated high dimensional variables. The compressed data set can now be divided into training set and test set. To classify if the driver is fatigued or not fatigued, we use Support Vector Classifier. This can also be correlated from the fact that Support Vector Classifier is highly efficient in working with the high dimensional feature vector. It is also very flexible in dealing with linearly as well as nonlinearly separable data sets.

SVM is a supervised learning method used for classification and regression. SVM is also referred to as Maximum Margin Classifier because it can maximize the geometric margin and minimize the emperical classification error simultaneoulsy. During classification SVM creates a maximal separating hyperplane. Now, two parallel hyperplanes are constructed on each side of hyperplane that separates the data. Here an assumption is made that larger the distance between the two parallel planes better is the generalization error of classifier [22].

SVM was used to implement this problem because of, binary nature of classification problem and the efficiency of SVM in working with high dimensional data and its flexibility in working with both linearly and nonlinearly separable data sets [23].

Unlike other classification algorithms, SVM can be used in both linear and non-linear ways with the use of a kernel. In cases when we have a limited set of points in many dimensions SVM tends to be very efficient because it can find linear separation in the data.

SVM is also eliminates the drawbacks of outliers as it uses only relevant points to find a linear separation, also called support vectors.

Now, we can train the classifier on the training set and can check its accuracy using test set. The test set and training set are completely different i.e. no two instances in both data sets are same. We do this so that our SVC model is always tested on a data which it has not seen before. This strategy provides a better picture of the generalized functioning of SVC. After the SVC model is trained we can calculate the prediction accuracy of our model using cross validation. In cross-validation, we perform a number of iterations, wherein each iteration we give a different data subset to test set from the previously allotted set. Finally, the classifier would return 1 if the driver is found to be in fatigue state or return -1 if the driver is found to be in an alert state to the alert unit.

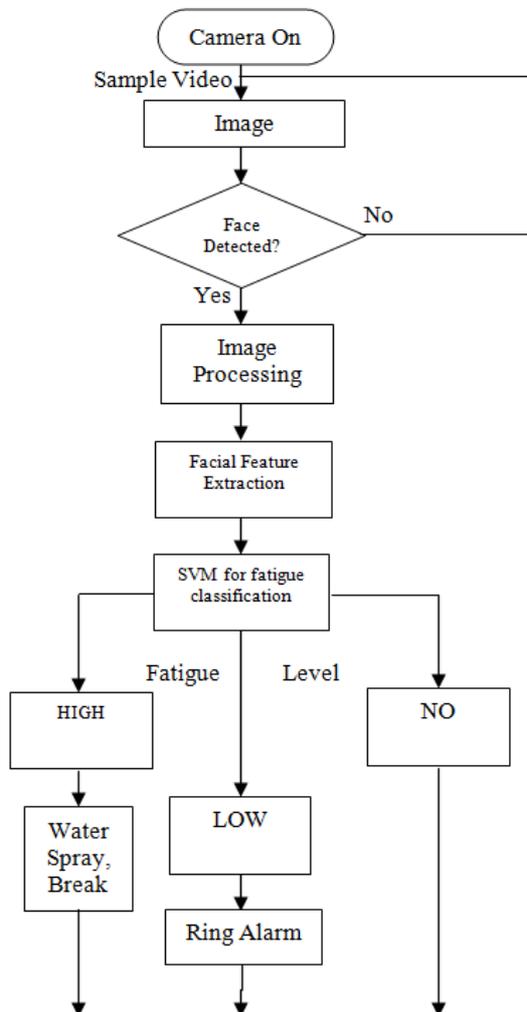

Figure 2. Flow chart of proposed system.

### D. Alert Unit

The modeling of the alert unit is done on the format (r, t), {r>=0, t>=0} where r is the resultant value and t is the time. The whole model depends on the output value of the fatigue detection unit. The Support Vector Machine (SVM) classifies the image, whether they are fatigued or not fatigued. The representation of the classification output is either a +1 or -1 and this number is used by the alert unit in the following way. Specifically, the classification output is the input of a running sum that adds the consecutive output values and has a minimum value up to zero. This unit employs two threshold level, first one is to detect whether there is no or low fatigue level. The second threshold value is to mark the difference between low and high fatigue level. Whenever the resultant value goes above a specified threshold, it implies that fatigue is detected and the alert system goes active and takes action according to the detected fatigue level. For low fatigue level, the alarm rings for 10sec and after 10sec the system again checks the resultant value and performs the alert process accordingly. For high fatigue level, we can use more effective accident preventive measures like an automatic reduce the speed and ultimately stop the vehicle and, or water spray. This system can be used to track the fatigue level of a driver and detect the sleep onset with a safe margin.

### V. DISCUSSION AND CONCLUSION

Our proposed system can be ahighly efficient system to monitor fatigue level in a driver and can dominate over disadvantages from previously developed methods by using both eye and mouth feature set and also with a vast pool of data. This system requires only a camera to monitor the driver's face, therefore reducing its hardware cost.

This system works only on the required part of face image i.e. eyes and mouth, rejects the rest. This step decreases the unnecessary features in the feature set. Eye and mouth detection is less accurate than the face detection, that's why we use the face detection to get the eye and mouth image part of the driver. It makes the system optimized in the context of time and accuracy. The division of system alert unit into three units is an efficient way to alert the driver. It works in such a way that the driver is not subjected to sudden attack, which may lead to an accident.

ACKNOWLEDGMENT

We are thankful to Dr. S. K. Raghuwanshi, Department of Electronics Engineering, IIT(ISM), Dhanbad for his valuable support and guidance.